%% file: main.tex
\documentclass[runningheads]{llncs}
\usepackage{graphicx}
\usepackage{tikz}
\usepackage{comment}
\usepackage{amsmath,amssymb} % define this before the line numbering.
\usepackage{color}

\usepackage[accsupp]{axessibility}  % Improves PDF readability for those with disabilities.

\usepackage[pagebackref,breaklinks,colorlinks]{hyperref}
\usepackage{subfigure} 
\usepackage{booktabs}
\usepackage{multicol}
\usepackage{multirow}
\usepackage{algorithm}
\usepackage{algorithmic}

\newcommand{\qileft}{[\kern-0.15em[}
\newcommand{\qiLeft}{\left[\kern-0.35em\left[}
\newcommand{\qiright}{]\kern-0.15em]}
\newcommand{\qiRight}{\right]\kern-0.35em\right]}

\newcommand{\R}{\mathbb{R}}

\newcommand{\T}{\mathcal{T}}

\newcommand{\A}{\mathcal{A}}  
\newcommand{\Y}{\mathcal{Y}}  
\newcommand{\X}{\mathcal{X}}

\newcommand{\mP}{\mathcal{P}}
\newcommand{\B}{\mathcal{B}}
\newcommand{\D}{\mathcal{D}}

\newcommand{\mL}{\mathcal{L}}
\newcommand {\bE}{\mathbb{E}} 
\newcommand {\y}{\boldsymbol{y}}
\newcommand {\x}{\boldsymbol{x}}
\newcommand {\z}{\boldsymbol{z}}

\newcommand {\mo}{\boldsymbol{o}}

\newcommand{\xmark}{\textbf{\textendash}}
\usepackage{array,booktabs,calc}  %  table plus function, like m
\usepackage{arydshln} % dashline
\newcolumntype{M}[1]{>{\centering\arraybackslash}m{#1}} % center table

\usepackage{enumitem}
\usepackage{balance}
\usepackage{color}

\definecolor{brightgreen}{rgb}{0.4, 1.0, 0.0}
\definecolor{chartreuse}{rgb}{0.5, 1.0, 0.0}
\definecolor{gain}{rgb}{0.224, 0.710, 0.290}
\newcommand{\sd}[1]{\scriptsize{$\pm#1$}}

\usepackage[capitalize]{cleveref}
\crefname{section}{Sec.}{Secs.}
\Crefname{section}{Section}{Sections}
\Crefname{table}{Table}{Tables}
\crefname{table}{Tab.}{Tabs.}

\begin{document}
% \renewcommand\thelinenumber{\color[rgb]{0.2,0.5,0.8}\normalfont\sffamily\scriptsize\arabic{linenumber}\color[rgb]{0,0,0}}
% \renewcommand\makeLineNumber {\hss\thelinenumber\ \hspace{6mm} \rlap{\hskip\textwidth\ \hspace{6.5mm}\thelinenumber}}
% \linenumbers
\pagestyle{headings}
\mainmatter
\def\ECCVSubNumber{1940}  % Insert your submission number here

\title{Online Continual Learning with Contrastive Vision Transformer} % Replace with your title

% INITIAL SUBMISSION 
\begin{comment}
\titlerunning{ECCV-22 submission ID \ECCVSubNumber} 
\authorrunning{ECCV-22 submission ID \ECCVSubNumber} 
\author{Anonymous ECCV submission}
\institute{Paper ID \ECCVSubNumber}
\end{comment}
%******************

% CAMERA READY SUBMISSION
% \begin{comment}
\titlerunning{Online Continual Learning with Contrastive Vision Transformer}
% If the paper title is too long for the running head, you can set
% an abbreviated paper title here

\author{Zhen Wang
\and
Liu Liu
\and
Yajing Kong
\and
Jiaxian Guo
\and
Dacheng Tao
}
\authorrunning{Z. Wang et al.}

\institute{The University of Sydney, Australia \\
\email{\{zwan4121,liu.liu1,ykon9947,jguo5934,dacheng.tao\}@sydney.edu.au}}
% \end{comment}
%******************
\maketitle

\begin{abstract}
Online continual learning (online CL) studies the problem of learning sequential tasks from an online data stream without task boundaries, aiming to adapt to new data while alleviating catastrophic forgetting on the past tasks.
This paper proposes a framework Contrastive Vision Transformer (CVT), which designs a focal contrastive learning strategy based on a transformer architecture, to achieve a better stability-plasticity trade-off for online CL.
Specifically, we design a new external attention mechanism for online CL that implicitly captures previous tasks' information.
Besides, CVT contains learnable focuses for each class, which could accumulate the knowledge of previous classes to alleviate forgetting.
Based on the learnable focuses, we design a focal contrastive loss to rebalance contrastive learning between new and past classes and consolidate previously learned representations.
Moreover, CVT contains a dual-classifier structure for decoupling learning current classes and balancing all observed classes.
The extensive experimental results show that our approach achieves state-of-the-art performance with even fewer parameters on online CL benchmarks and effectively alleviates the catastrophic forgetting.

\end{abstract}

\input{text/1intro}

\input{text/2related}
\input{text/3method}

\input{text/4exp}

\input{text/5conclu}

% Papers accepted for the conference will be allocated 14 pages (plus additional pages for references) in the proceedings. 

% ---- Bibliography ----
%
% BibTeX users should specify bibliography style 'splncs04'.
% References will then be sorted and formatted in the correct style.
%
\bibliographystyle{splncs04}
\bibliography{egbib}

\end{document}

%% file: text/1intro.tex
\section{Introduction}
\label{Osec:intro}

One of the major challenges in research on artificial neural networks is developing the ability to accumulate knowledge over time from a non-stationary data stream~\cite{DL,ICML20_guo,domainIL}.
Although most successful deep learning techniques can achieve excellent results on pre-collected and fixed datasets, they are incapable of adapting their behavior to the non-stationary environment over time~\cite{PAMI_survey,domainIL2}.
When streaming data comes in continuously, training on the new data can severely interfere with the model's previously learned knowledge of past data, resulting in a drastic drop in performance on the previous tasks. 
This phenomenon is known as \textit{catastrophic forgetting} or \textit{catastrophic interference}~\cite{catastrophic,connectionist}.
Continual learning (also known as lifelong learning or incremental learning)~\cite{iCaRL,ER,CVPR22_LVT,domainIL2,ICML_wokrshop21,zhu2020iexpressnet,PAMI_survey} aims to solve this problem by maintaining and accumulating the acquired knowledge over time from a stream of non-stationary data.

% Stability-plasticity dilemma arises in continual learning since it requires neural networks to be both stable to prevent forgetting as well as plastic to learn new concepts~\cite{dilemma,adaptive}. 

Continual learning requires neural networks to be both stable to prevent forgetting as well as plastic to learn new concepts, which is referred to as the  \textit{stability-plasticity} dilemma~\cite{dilemma,adaptive}.
The early works have focused on the Task-aware protocol, which selects the corresponding classifier using oracle knowledge of the task identity at inference time~\cite{MAS,LwF,task-IL-HAT,rectification_expanding,PAMI_survey}.
For example, regularization-based methods penalize the changes of important parameters when learning new tasks and typically assign a separate classifier for each task~\cite{RW,EWC,oEWC,xi2022single,SI}.
Recent studies have focused on a more practical Task-free protocol, which evaluates the network on all classes observed during training without requiring the identity of the task~\cite{expert-gate,AAAI22_CL,FDR,zhu2020iexpressnet,rebalancing,semantic_IL,HAL}.
Among them, rehearsal-based methods that store a small set of seen examples in a limited memory for replaying have demonstrated promising results~\cite{GSS,iCaRL,DER,RehearsalICCV21}.
This paper focuses on a more realistic and challenging setting: online continual learning (online CL)~\cite{online20,online_review,ASER}, where the model learns a sequence of classification tasks with a single pass over the data and without task boundaries. There are also Task-free and Task-aware protocols in online CL.

Inspired by recent breakthroughs in self-supervised learning~\cite{chen2020simple,unsupervised_cont,CVPR22_GUO,tian2019contrastive}, we find that the knowledge learned by supervised contrastive learning~\cite{contrastive06,SCL,SCL_pretrain} reveals greater robustness and transferability.
The general and transferable knowledge is the essence of what online CL seeks to explore, which could effectively help mitigate forgetting.
Unfortunately, it is challenging to employ contrastive learning in the continual setting. There are two main issues as follows: 1) Contrastive learning requires informative negative samples to learn and distinguish different clusters. However, in online CL, previous task data is unavailable or very limited, which causes severe imbalanced contrast between past and new classes. 2) The contrastively learned knowledge may suffer from 
forgetting since the distribution of the data stream is continually changing.

\begin{figure*}[tb]
    \centering
    \includegraphics[width=\textwidth]{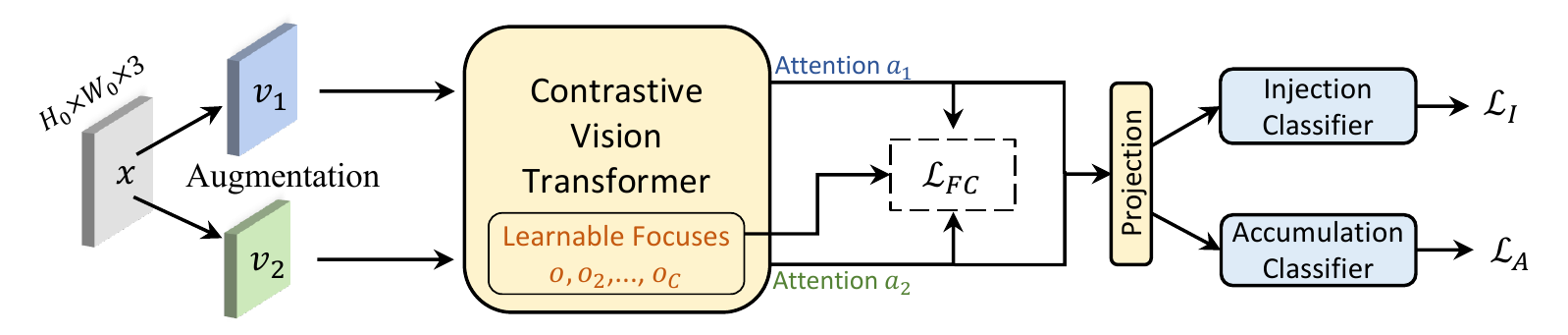}
    \caption{The overall framework of Contrastive Vision Transformer (CVT).}
    \label{Ofig:framework}
\end{figure*}

Considering the superior modeling capability of Vision Transformers~\cite{vit} recently demonstrated on computer vision tasks~\cite{transferable,CvT,levit,T2T,CoAtNet}, we take the utilization of the potential of attention mechanism~\cite{attention} to develop online CL.
Overall, we strategically integrate contrastive learning and transformer to model the online data stream. We propose a novel framework, Contrastive Vision Transformer (CVT), to alleviate the forgetting problem and tackle the above imbalance issue of contrastive learning in online CL.
An overview of the framework is illustrated in \cref{Ofig:framework}. Specifically, we newly design an effective and efficient transformer architecture with external attention to implicitly capture previous tasks' information and reduce the number of parameters.
CVT contains learnable focuses for each class, which could accumulate the knowledge of previous classes to alleviate forgetting.
Based on the learnable focuses, we design a focal contrastive loss at the attention level to rebalance contrastive learning between new and past classes and improve the inter-class distinction and intra-class aggregation.
Moreover, CVT contains a dual-classifier structure: an \textit{injection classifier} is used to inject representation of stream data into the model, mitigating interference with previous knowledge; and an \textit{accumulation classifier} focuses on integrating the previous and new knowledge in a balanced manner.

We systematically compare state-of-the-art and well-established methods for the online CL problem in both the Task-free and Task-aware protocols.
Experimental results show that the CVT framework significantly outperforms other approaches in terms of accuracy and forgetting, even with fewer parameters. Ablation study validates each component of the proposed framework.

\noindent The main contributions of this paper are three-fold: 
\begin{itemize}
    \item We propose a novel framework Contrastive Vision Transformer (CVT) to achieve a better stability-plasticity trade-off for online CL. CVT contains class-wise learnable focuses, which can accumulate the knowledge of previous classes to alleviate forgetting.
    \item We design a focal contrastive loss to rebalance contrastive learning between new and past classes and learn more robust representations. 
    \item The extensive experimental results show that CVT achieves state-of-the-art performance with even fewer parameters on online continual learning benchmarks.
\end{itemize}

%% file: text/2related.tex
\section{Related Works}
\label{Osec:related}

\subsection{Continual Learning Methods}

Continual learning (CL) methods have been developed to alleviate catastrophic forgetting in neural networks. 
These methods can be divided into three main categories: expansion-based, regularization-based, and rehearsal-based methods.

As new tasks arrive, expansion-based methods dynamically expand networks and keep sub-networks related to previous tasks fixed~\cite{rectification_expanding,DER_cvpr21,conditional,randomPath,xi2020bounding,Grow,TNNLS_SIN,piggyback}. 
However, most expansion-based methods require task identity during inference in order to allocate distinct sets of parameters to distinct tasks.
Regularization-based methods limit the changes in important parameters during the learning of new tasks by estimating the importance of each network parameter for prior tasks~\cite{RW,EWC,MAS,oEWC,Curriculum,Bit-Level,SI}.
These works differ in how they compute the importance of network parameters.

Rehearsal-based methods~\cite{FDR,rebalancing,AAAI22_CL,Encoder-CL,CHEN2022365,Podnet,GEM,CVPR22_LVT,geodesic,SCR} alleviate catastrophic forgetting by replaying a subset of data of past tasks stored in limited buffer. 
iCaRL~\cite{iCaRL} trains a nearest-class-mean classifier while limiting the change of the representation in later tasks through a self-distillation loss.  
In addition to replaying past experiences, HAL~\cite{HAL} keeps predictions on some \textit{anchor} points by an additional objective. IExpressNet~\cite{zhu2020iexpressnet} introduces representative expression memory while employing a novel Center-expression-distilled loss and having shown satisfactory performance for the task of facial expression recognition.
DER++~\cite{DER} combines rehearsal with distillation loss~\cite{KD,ICML20_DSL,KDimproving} to retrain past experience and obtain state-of-the-art performance.
RM~\cite{rainbow} proposes an uncertainty-based sampling approach by using uncertainty and data augmentation to improve rehearsal.
The proposed method in this paper belongs to the rehearsal-based method.

\subsection{Online Continual Learning}
Online continual learning (online CL) is a more realistic~\cite{online_review,AAAI21_MFL,xi2020incentive,zhu2019physiological,ASER,GEM} and difficult setup~\cite{online20}, where the model learns from a non-i.i.d data stream online, without the help of task identifiers or task boundaries both at training and inference stages.
Online CL methods are mainly based on rehearsal.

Experience Replay (ER)~\cite{ER} employs reservoir sampling for memory management and jointly optimizes a network by mixing memory data with online stream data. ERT~\cite{er_tricks} improves ER by balanced sampling and bias control.
AGEM~\cite{A-GEM} and GEM~\cite{GEM} use episodic memory to compute past task gradients to constrain the online update step.
GSS~\cite{GSS} presents a gradient-based sampling to store diversified data for learning more information. 
ASER~\cite{ASER} is based on the Shapley Value theory to improve the memory buffer update and sampling.
CLS~\cite{FastSlow} uses two extra models to maintain long-term and short-term semantic memories for knowledge consolidation.
SCR~\cite{SCR} uses supervised contrastive loss for representation learning and employs the nearest class mean to classify.

\subsection{Vision Transformers}
Transformer model is firstly applied into machine translation tasks~\cite{attention}, and then, Transformers have become the state-of-the-art models for most natural language processing tasks~\cite{bert,language_transformer,DUAN202228,translation,video}. 
Attention modules are the core components of transformers, which aggregate information from the entire input sequence.
Recently, Vision Transformer (ViT)~\cite{vit} is proposed to makes Transformer architecture scalable for image classification when the data is large enough.
Since then, a lot of effort has been dedicated to improving Vision Transformers' data efficiency and model efficiency~\cite{T2T,T_survey,ViTCL,T_survey2}, where an effective direction is to strategically integrate properties of convolution into the Transformer architecture~\cite{scaling,ViTCL,Swin,CvT,levit,zhao2021survey,DyTox,CoAtNet}.
CoAT~\cite{CoAT} proposes a conv-attention module that realizes relative position embeddings with convolutions.
LeViT~\cite{levit} builds pyramid attention structure with pooling to learn convolutional-like features. 
By leveraging sequence pooling and convolutions, CCT~\cite{CCT} eliminates the need for class tokens and positional embeddings.

Nevertheless, current vision transformers may not be applicable to modeling the online data stream, and existing continual learning algorithms developed for CNNs may not be ideal for vision transformers as well.
To this end, we propose a lightweight Contrastive Vision Transformer (CVT) with a focal contrastive loss for online continual learning and achieve better performance than other transformers and CNN baselines.

%% file: text/3method.tex
\section{Preliminary}
\label{Osec:Preliminary}

\subsubsection{Problem Setup.}
Formally, an online continual learning problem is split in a sequence of $T$ supervised learning tasks $\T_t$, $t\in\{1,...,T \}$, where $T$ is the total number of tasks. Let $\mathcal{D} = \{\mathcal{D}_1, ..., \mathcal{D}_T\}$ be the corresponding online data stream, where $\mathcal{D}_t$ is the dataset of task $\mathcal{T}_t$. 
For task $\T_t$, input samples $x\in\X_t$ and the corresponding ground truth labels $y\in\Y_t$ are drawn from the i.i.d. distribution $\D_t$.
% The training data comes gradually in a stream and is only trained incrementally in a single iteration. 
A mini-batch of training data $\B$ from $\D$ comes gradually in an online stream (each sample is seen only once).
Besides, a limited memory buffer $\mathcal{M}$ saves a small set of training data of seen tasks.
The model is trained on $\B \cup \mathcal{M}$ at each iteration.
At task $\T_t$, The label space of the model is all observed classes $\cup_{i=1}^t \mathcal{Y}_i$,
and the model is expected to predict well on all classes at the inference stage.

\subsubsection{Supervised Contrastive Learning (SCL). }
SCL~\cite{SCL,SCL_pretrain,chen2020simple} aims to push the representation of samples with different classes farther apart while tightly clustering representation of samples with the same class.
% We formulate SCL in the following.
Suppose that the classification model can be decomposed into two components: an encoder $f$ and a classifier $w$. 
Encoder $f$ maps an image sample $\x$ to a vectorial embedding (representation) $\z = f(\x)$. Classifier $w$ maps the representation $\z$ to a classification vector $\hat{\y} = w(\z)$.
Without training $w$, SCL focuses on training $f$ as follows: given a batch of $b$ samples $\mathcal{B} = \{(\x_k, y_k)\}_{k=1}^b$, SCL first generates an augmented batch $\widetilde{\mathcal{B}} = \{(\tilde{\x}_k, \tilde{y}_k)\}_{k=1}^{2b}$ by making two random augmentations of $\mathcal{B}$, with $y_k = \tilde{y}_{2k-1} = \tilde{y}_{2k}$.
The SCL loss takes the following form:

\begin{equation}
    \mathcal{L}_{\text {SCL}}=\sum_{i \in \mathcal{I}} \frac{-1}{|\mP(i)|} \sum_{\z_p \in \mP(i)} \log \frac{\exp \left({\z}_{i} \cdot {\z}_{p} / \tau\right)}{\sum_{\z_j \in \A(i)} \exp \left({\z}_{i} \cdot {\z}_{j} / \tau\right)}, 
\label{Oeq:scl}
\end{equation} 
where $\mathcal{I}$ represents the set of indices of $\widetilde{\B}$; $\mathcal{A}(i)\equiv\{z_i: i\in\mathcal{I}\backslash\{i\} \}$ is the set of representations of samples in $\widetilde{\B}$ except for that of $\x_i$; $\mathcal{P}(i): \equiv\left\{\z_p \in \A(i): \tilde{y}_{p}=\tilde{y}_{i}\right\}$ is the set of representations of positive samples (i.e., samples with the same labels) with respect to the anchor $\tilde{\x}_i$; $\tau \in \R^{+}$ is a temperature hyperparameter; $|\mP(i)|$ is its cardinality.

\begin{figure*}[t]
    \centering
    \includegraphics[width=\textwidth]{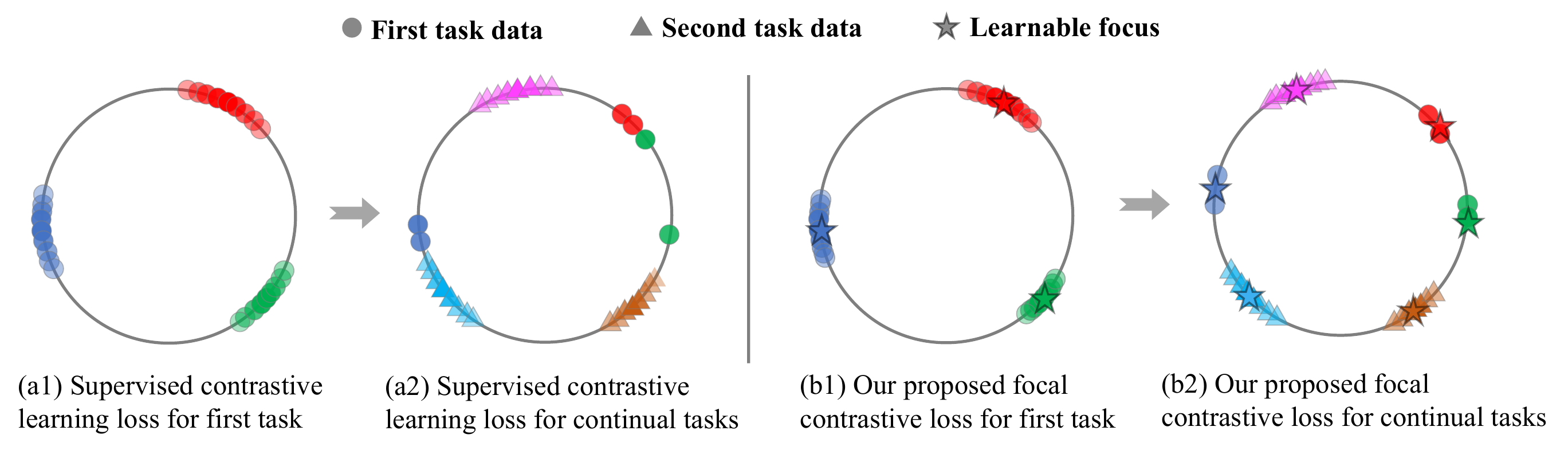}
    \caption{Continually arriving samples are clustered on the hypersphere by contrastive learning (for visualization clarity, we use a 2D representation space). 
    Different colors represent different classes. Supervised contrastive learning loss on online stream data fails to obtain good inter-class distinction and intra-class aggregation caused by the class imbalance.
    Focal contrastive loss effectively mitigates class imbalance in online CL and accumulates class-wise knowledge by the learnable \textit{focuses}.}
    \label{Ofig:SCL}
\end{figure*}

Although SCL could learn the transferable  representation to help prevent forgetting in online CL, SCL will face new challenges: 1) previous task data is unavailable or very limited due to the streaming fashion, which causes severe imbalanced contrast between past and new classes;
2) the contrastively learned knowledge may suffer from forgetting since the distribution of the data stream is continually changing. For visualization
clarity we use a 2D feature space. As illustrated in \cref{Ofig:SCL}(a1) and \cref{Ofig:SCL}(a2), the imbalanced data stream in online CL makes the representation of previous tasks drift and difficult to be accurately clustered by SCL.

\section{Methodology}
\label{Osec:method}

To alleviate the forgetting problem in online continue learning, we propose a framework Contrastive Vision Transformer (CVT), which designs a new focal contrastive learning strategy based on the transformer architecture. 
An overview of the framework is depicted in \cref{Ofig:framework}. 
CVT plays the strengths of the attention mechanism in online CL, which design an effective transformer architecture with external attention.
We tackle the imbalance issue of SCL in online CL by proposing a focal contrastive loss at the attention level.
The learnable focuses in CVT can accumulate class-specific knowledge to alleviate the forgetting of previous tasks.
Besides, a dual-classifier is used to decouple learning current classes and balancing all seen classes, improving the stability-plasticity trade-off.

In the following, we will go through the description of CVT in terms of both model architecture (\cref{Osubsec:archit}) and focal contrastive continual learning strategy (\cref{Osubsec:strategy}), respectively.

\subsection{Model Architecture}
\label{Osubsec:archit}
\begin{figure*}[t]
    \centering
    \includegraphics[width=\textwidth]{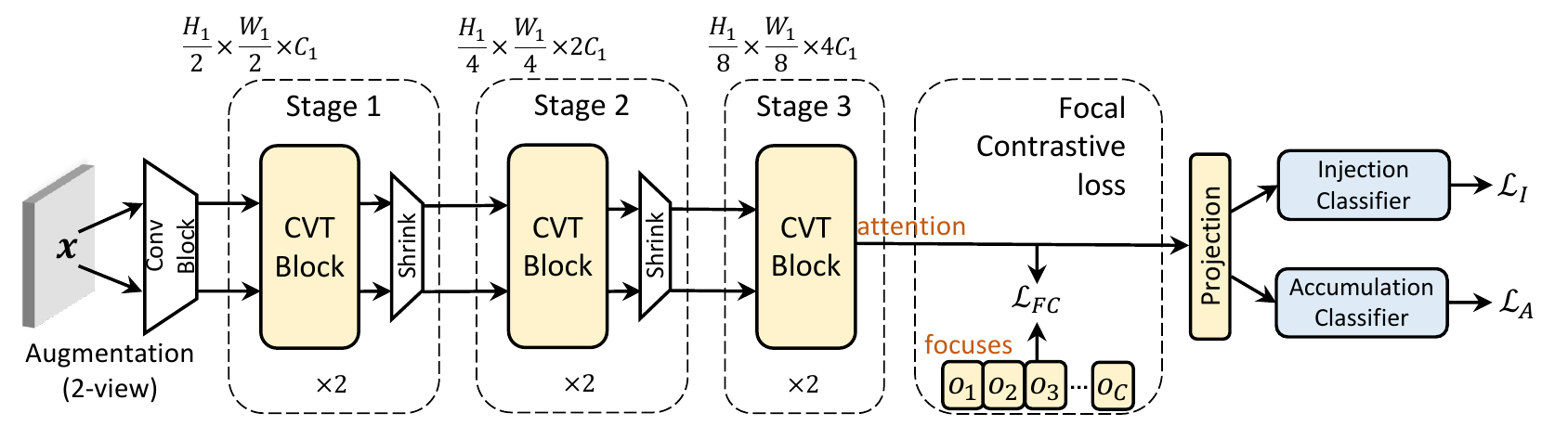}
    \caption{The architecture of Contrastive Vision Transformer (CVT). CVT architecture is composed of stacked  transformer blocks after a simple convolutional block. Shrink module performs downsampling to reduce the resolution of the activation maps and increase their number of channels between CVT stages. Focuses are a set of learnable attention vectors. After a projection layer, two classifiers serve for knowledge injection and accumulation, respectively.}
    \label{Ofig:architecture}
\end{figure*}

\cref{Ofig:architecture} illustrates the CVT architecture. 
The major contributing components in the architecture include 1) \textit{external attention}, which implicitly captures previous tasks' information and reduces the number of parameters, and 2) \textit{learnable focuses}, which could maintain and accumulate the knowledge of previous classes.

\subsubsection{External Attention.}
CVT plays the strengths of the attention mechanism in online CL.
Unlike the vanilla self-attention in vision transformers~\cite{vit,levit} that derives the attention map by computing similarity between self-queries and self-keys~\cite{attention}, we introduce an external attention mechanism~\cite{CVPR22_LVT} to obtain the attention map by computing the affinities between self-queries and a learnable external key $K_W$ with an attention bias $B$, which implicitly injects previous task information in the attention mechanism.
% and reduces parameters of the model. 
Moreover, the proposed architecture can save the number of parameters compared to self-attention.

Let the input tensor be $X$, we apply linear transformation with weights $W_q$ and $ W_v$ to get the vanilla self-query $Q_X$$=$$W_q X$ and self-value $V_X$$=$$W_v X$, respectively. 
We employ a linear layer $K_W$ to replace the input-depended self-key, and explicitly add a learnable attention bias $B$ to attention maps. 
Consider $H$ attention heads, which are uniformly split into $H$ segments $Q_X^h, K_W^h, V_X^h$, and $B^h$.
The external attention mechanism computes the head-specific attention map $A^h$ and concatenates the multi-head attention as follows:
% aggregate the token value features as follows:
\begin{equation}
\begin{aligned}
    A^h \!=\! \text{Softmax}\left(\frac{\text{Norm}(Q^{h}_X (K_W^{h})^\top) \!+\! B^h}{\sqrt{d/H}} \right), 
   \quad X_{out}^h = A^h V_X^h , \quad h = 1,...,H,
\end{aligned}
\end{equation}
where Norm() denotes batch normalization; $d$ is the dimension of the key.

\subsubsection{Learnable Focuses.}
CVT contains learnable \textit{focuses} for each class, which could maintain and accumulate the knowledge of previous classes to alleviate forgetting for online CL. 
The class-wise learnable focuses $\mathcal{O} = \{\mo_1, \mo_2,..., \mo_C\}$ is a set of learnable attention vectors, as shown in \cref{Ofig:architecture}, where a focus $\mo_c$ corresponds to class $c$, and $C$ is the number of the seen classes.
The size of the learnable focuses is negligible in relation to the overall model.

When a new class $c$ appears in the data stream, the corresponding focus $\mo_c$ to class $c$ starts to participate in the training (refer to \cref{Oeq:FCloss}). Even if class $c$ no longer appears in the data stream afterward, the focus $\mo_c$ always participates in the online CL training and serves as a negative sample for the other classes, as illustrated in \cref{Ofig:SCL}(b1) and \cref{Ofig:SCL}(b2).
Thus, focuses $\mathcal{O}$ preserve and accumulate the previously learned class-specific knowledge and acts as a forgetting mitigation in online continual learning.

\subsection{Focal Contrastive Continual Learning}
\label{Osubsec:strategy}
We propose a rehearsal-based focal contrastive learning scheme to 1) tackle the imbalance issue of SCL in online CL and 2) accumulate class-specific knowledge, to alleviate the interference with previous tasks.
The learning scheme includes two losses: a focal contrastive loss and a dual-classifier loss, as following.

\subsubsection{Focal Contrastive Loss.} For learning representation continually, we propose a focal contrastive loss $\mathcal{L}_\text{FC}$ in online CL.
As mentioned in \cref{Osec:Preliminary}, during the training phase, the model observes a mini-batch $\B$ at a time sampled from task $\T_t$ in the data stream $\D$. An input batch for the model is composed of $\B$ and $\B_{\mathcal{M}}$ sampled from the memory buffer $\mathcal{M}$.
The input batch and its augmented view are encoded by CVT blocks to generate attention $\z$, as shown in \cref{Ofig:architecture}.  As mentioned previously, a set of class-wise learnable \textit{focuses} $\mathcal{O} = \{\mo_1, \mo_2,..., \mo_C\}$ is utilized by focal contrastive loss $\mathcal{L}_\text{FC}$, where a focus $\mo_c$ is a learnable attention vector for class $c$ and $C$ is the number of the seen classes.
The FC loss function is defined as:

\begin{equation}
\begin{gathered}
    \mathcal{L}_{\text{FC}}=\sum_{i \in \mathcal{I}} \frac{-1}{|\mP(i)\cup \mo_{\tilde{y}_i}|} \sum_{\z_p \in \mP(i)\cup \mo_{\tilde{y}_i}} \delta_{\z_p} \log  \frac{\exp \left({\z}_{i} \cdot {\z}_{p} / \tau\right)}{\sum_{j \in \A(i)\cup\mathcal{O}} \exp \left({\z}_{i} \cdot {\z}_{j} / \tau\right)}, 
\label{Oeq:FCloss}
\end{gathered}
\end{equation}

where

\begin{align}
    \delta_{\z_p}  = 
    \begin{cases}\mu,  \qquad \z_p \in \mathcal{O} \\
    1.0,  \ \quad  \z_p \in \mP(i) \end{cases},
\end{align}
and $\mu$ is the weight of focuses. We set $\mu > 1$ to make focuses play a more important role in contrastive learning; $\mP(i)$ and $\A(i)$ are the same with supervised contrastive learning in \cref{Oeq:scl}; $\tau \in \R^{+}$ is a temperature hyperparameter.

The benefits of using focal contrastive loss are two-fold. First, it alleviates the imbalance issue in online CL by employing the class-wise focuses.
Second, it accumulates the previous knowledge by the learnable focuses which will continually serve as the prototypes of classes to maintain class-specific information.
With a proposed focal contrastive loss $\mL_\text{FC}$ in training, CVT rebalances contrastive learning between new and past classes and improve the inter-class distinction and intra-class aggregation, as illustrated in \cref{Ofig:SCL}. In \cref{Osubsec:ablation}, we empirically observe that $\mL_\text{FC}$ outperforms the original $\mL_\text{SCL}$ and boost the online CL.

\subsubsection{Dual-classifier Loss.}
We propose a dual-classifier structure to decouple learning current classes and balancing all seen classes, which contains an \textit{injection classifier} to inject new task representation into the model, alleviating interference to previously learned knowledge, and an \textit{accumulation classifier} to integrate past and new knowledge in a balanced manner.

Let $g(\x)$ be the representation of a sample $\x$ outputted from the Projection of CVT before the classifier. 
When new data stream batch arrives, we utilize the output from an independent injection classifier to compute a classification loss:
\begin{equation}
\label{Oeq:injectionLoss}
    \mathcal{L}_{I} = {\sum_{(\x,y)\in \B} \ell(y, f_{I}(g(\x)))},
\end{equation}
% \begin{equation}
% \label{Oeq:injectionLoss}
%     \mathcal{L}_{I} = {\bE_{(x, y) \sim \D_t}\big[\ell(y, f_{I}(g(x)))\big]},
% \end{equation}
where $f_{I}$ denotes the injection classifier and $\ell$ adopts a cross-entropy loss. 
$f_{I}$ is only trained on stream data and does not participate in the inference stage.

Besides, we employ an accumulation classifier to focus on improving the stability-plasticity trade-off by integrating previous and new knowledge in a balanced manner. 
The accumulation classifier is used at the inference stage for outputting the prediction. 
Rehearsing the limited memory data during learning new tasks is a crucial way to maintain previous knowledge. We can replay the exemplars stored in the memory buffer with their ground truth labels. In addition, the accumulation classifier also needs a supervised signal from current task data.
Therefore, we give the accumulation classifier loss:
\begin{equation}
\label{Teq:lifelongloss}
    \mathcal{L}_{A} = \alpha\bE_{(\x',y') \sim \mathcal{M}}\big[ \ell(y', f_{A}(g(\x')))\big] + \beta{\sum_{(\x,y)\in \B}  \ell(y, f_{A}(g(\x)))} , 
\end{equation}
where $f_{A}$ denotes the accumulation classifier; $\alpha$ and $\beta$ are the coefficients balancing knowledge consolidation. We approximate the expectation by computing gradients on batches sampled from the memory buffer.

Overall, the total loss used in CVT is the sum of Eq.~\eqref{Oeq:FCloss}, Eq.~\eqref{Oeq:injectionLoss}, and Eq.~\eqref{Teq:lifelongloss}: 
% $\mathcal{L} =  \mL_A + \mL_{I} + \gamma\mL_\text{FC}$, 
\begin{equation}
    \mathcal{L} =  \mL_A + \mL_{I} + \gamma\mL_\text{FC},
\end{equation}
where $\gamma$ is the coefficient balancing $\mL_\text{FC}$.
After updating the whole CVT model on a mini-batch $\B$, the memory buffer $\mathcal{M}$ will be updated with $\B$ by Reservoir sampling~\cite{reservoir,DER}.

%% file: text/4exp.tex
\section{Experiment}
\label{Osec:exp}

\begin{table*}[t]
\centering
\resizebox{\columnwidth}{!}{%
\setlength{\tabcolsep}{5pt}
\begin{tabular}{@{}cllcccc@{}}
\toprule
\multirow{2}{*}{\begin{tabular}[c]{@{}c@{}}\textbf{Memory}\\ \textbf{Buffer}\end{tabular}} & \multirow{2}{*}{\textbf{Method}} & \multirow{2}{*}{$\#$\textbf{Paras}} & \multicolumn{2}{c}{10 \textbf{splits}} & \multicolumn{2}{c}{20 \textbf{splits}} \\
 &  &  & \textit{Task-free} & \textit{Task-aware} & \textit{Task-free} & \textit{Task-aware} \\ \midrule
 {\xmark} & SGD &11.2  &5.77\sd{0.35}  &37.44\sd{1.83}  &3.53\sd{0.12}  &36.81\sd{2.63} \\
 \midrule
\multirow{11}{*}{500} & ER~\cite{ER} &11.2  &15.59\sd{1.20}  &59.72\sd{0.83}  &12.51\sd{0.77}  &62.72\sd{1.70}  \\
 & GEM~\cite{GEM} &11.2  &14.34\sd{1.63}  &50.39\sd{0.28}  &5.98\sd{0.33}  &57.15\sd{1.77} \\
 & AGEM~\cite{A-GEM} &11.2  &6.35\sd{0.13}  &39.18\sd{0.27}  &3.62\sd{0.08}  &39.55\sd{0.17}\\
 & iCaRL~\cite{iCaRL} &11.2   &15.18\sd{0.31}  &48.95\sd{0.33}  &12.79\sd{0.26}  &60.53\sd{0.38}\\
 & FDR~\cite{FDR} &11.2  &5.97\sd{0.20}  &32.18\sd{1.60}  &3.60\sd{0.07}  &39.98\sd{1.32}\\
 & GSS~\cite{GSS} &11.2  &10.91\sd{0.36}  &59.10\sd{0.26}  &6.33\sd{0.30}  &62.80\sd{2.31}\\
 & DER++~\cite{DER} &11.2  &15.72\sd{1.29}  &54.45\sd{1.59}  &11.29\sd{0.25}  &63.62\sd{1.05} \\
  & HAL~\cite{HAL}  &22.4 &10.51\sd{0.63}  &33.70\sd{1.37}  &7.09\sd{0.39}  &52.89\sd{2.36}\\
 & ERT~\cite{er_tricks} &11.2  &16.28\sd{0.73} &60.11\sd{2.18}  &17.92\sd{0.42}   &68.08\sd{0.37}  
 \\
 & ASER$_\mu$~\cite{ASER} &11.2  &12.42\sd{0.68} &53.77\sd{1.30} &9.63\sd{0.51}  &58.91\sd{1.62}
 \\
 &RM~\cite{rainbow} &11.2 &14.32\sd{0.49}  &58.76\sd{1.82}  &13.73\sd{0.60}  &64.73\sd{2.01}\\
 &CLS~\cite{FastSlow} &33.8  &15.06\sd{0.57}  &59.82\sd{1.24}  &14.84\sd{0.83}  &65.74\sd{1.94}  
 \\
 & \textbf{CVT (ours)} &\textbf{8.9} &\textbf{24.45}\sd{0.62}  &\textbf{62.52}\sd{1.43}
 &\textbf{21.81}\sd{0.52}
 &\textbf{71.23}\sd{1.40}
    \\
\midrule
\multirow{11}{*}{1000} & ER~\cite{ER} &11.2  &20.41\sd{1.46}  &63.39\sd{1.37}   &17.02\sd{1.63}  &68.52\sd{1.09}   \\
 & GEM~\cite{GEM} &11.2  &16.49\sd{1.08}  &52.30\sd{0.51}  &8.40\sd{1.05}  &62.59\sd{1.89} \\
 & AGEM~\cite{A-GEM} &11.2  &6.57\sd{0.12}  &40.38\sd{0.30}  &3.74\sd{0.05}  &42.39\sd{0.42}\\
 & iCaRL~\cite{iCaRL} &11.2   &16.31\sd{0.26}  &50.49\sd{0.18}  &13.03\sd{0.26}  &61.13\sd{0.20}\\
 & FDR~\cite{FDR} &11.2  &6.58\sd{0.31}  &36.99\sd{0.45}  &3.72\sd{0.04}  &42.45\sd{0.39}\\
 & GSS~\cite{GSS} &11.2  &12.38\sd{0.59}  &60.75\sd{0.28}  &7.40\sd{0.23}  &66.06\sd{1.25}\\
 & DER++~\cite{DER} &11.2  &21.27\sd{1.69}  &61.80\sd{1.24}  &13.42\sd{0.50}  &71.26\sd{0.61} \\
  & HAL~\cite{HAL}  &22.4 &11.81\sd{0.79}  &39.67\sd{2.63}  &13.14\sd{0.72}  &60.03\sd{1.20}\\
 & ERT~\cite{er_tricks} &11.2  &23.43\sd{0.58} &62.25\sd{1.33}  &24.58\sd{0.36}   &72.61\sd{0.67} \\
  & ASER$_\mu$~\cite{ASER} &11.2  &14.38\sd{0.43} &58.91\sd{1.76} &12.79\sd{0.60} &62.47\sd{1.82} 
  \\
 &RM~\cite{rainbow} &11.2 &22.41\sd{1.28}  &61.82\sd{2.13}  &18.91\sd{1.15}  &67.30\sd{1.34}
 \\
  &CLS~\cite{FastSlow} &33.8  &19.73\sd{1.17}  &62.54\sd{1.52}  &17.06\sd{1.47}  &70.08\sd{0.85}  
  \\
 & \textbf{CVT (ours)} &\textbf{8.9} &\textbf{28.83}\sd{0.86}  
 &\textbf{65.86}\sd{1.24}
 &\textbf{28.15}\sd{0.51}
 &\textbf{75.76}\sd{0.93}  
\\ \bottomrule
\end{tabular}  
}
\caption{Results (overall accuracy \%) on CIFAR100 benchmark which is averaged over five runs. \#Paras means the number of parameters in the model, which is counted by million.}
\label{Otab:cifar100}
\end{table*}

\subsection{Experimental Setup and Implementation}
\label{Osubsec:setup}

We consider a strict evaluation setting~\cite{domainIL,domainIL2} for online continual learning, including task-aware protocol~\cite{task-aware} and task-free protocol~\cite{task-free}.
For task-aware protocol~\cite{task-aware}, the task identities are required for each time evaluation. For task-free protocol~\cite{task-free}, the task identities are unavailable at inference time.
% as the harder scenario~\cite{domainIL2,domainIL}.

\noindent\textbf{Datasets.} 
Online continual learning benchmarks evaluate the capacity of an algorithm to learn on not independent and identically distributed (non-iid) data.
CIFAR-100~\cite{cifar} contains 100 classes and each class has 100 testing and 500 training color images.
TinyImageNet~\cite{tinyimgnet} consists 200 classes that include 100,000 images for training and 10,000 images for testing.
ImageNet100~\cite{iCaRL} contains 100 classes randomly chosen from ILSVRC~\cite{imagenet}, including about 120,000 images for training and 5,000 images for validation.

\noindent \textbf{Baselines.}
We compare CVT with state-of-the-art and well-established Online CL baselines, including 11 rehearsal-based methods (ER~\cite{ER}, GEM~\cite{GEM}, AGEM~\cite{A-GEM}, GSS~\cite{GSS}, FDR~\cite{FDR}, HAL~\cite{HAL}, ASER$_\mu$~\cite{ASER}, ERT~\cite{er_tricks}, RM~\cite{rainbow}, SCR~\cite{SCR}, and CLS~\cite{FastSlow}), 2 methods leveraging Knowledge Distillation (iCaRL~\cite{iCaRL} and DER++~\cite{DER}). Besides, we also compare vision transformers (ViT~\cite{vit}, LeViT~\cite{levit}, CoAT~\cite{CoAT}, and CCT~\cite{CCT}) with rehearsal strategy for continual learning. We also provide the results of simply performing \textit{SGD} without any countermeasure to alleviate forgetting.

\begin{table*}[t]
\centering
\small
\resizebox{\columnwidth}{!}{%{
\setlength{\tabcolsep}{3pt}
\begin{tabular}{@{}cl|lll|lllcc@{}}
\hline
\multirow{2}{*}{\begin{tabular}[c]{@{}c@{}}\textbf{Memory}\\ \textbf{Buffer}\end{tabular}} & \multirow{2}{*}{\textbf{Method}} & \multirow{2}{*}{$\#$\textbf{Paras}} & \multicolumn{2}{c|}{\textbf{TinyImageNet}} & \multirow{2}{*}{$\#$\textbf{Paras}} & \multicolumn{2}{c}{\textbf{ImageNet100}} \\
 &  &  &  \textit{Task-free} & \textit{Task-aware} & & \textit{Task-free} & \textit{Task-aware}  \\ \hline
{\xmark} & SGD &11.2 &4.54\sd{0.03}  &26.25\sd{0.16} &11.2 &3.98\sd{0.02} &19.05\sd{0.17} \\ \hline
\multirow{8}{*}{500} 
&ER~\cite{ER}  &11.2 &9.71\sd{0.18}  &42.76\sd{0.35} &11.2 &9.88\sd{0.46} &32.38\sd{0.63} \\
&AGEM~\cite{A-GEM}  &11.2  &4.63\sd{0.08}  &27.86\sd{0.13} &11.2 &3.38\sd{0.08}  &21.80\sd{0.15}\\
&iCaRL~\cite{iCaRL}  &11.2 &6.17\sd{1.03}  &27.22\sd{1.52} &11.2 &7.70\sd{0.32} &20.45\sd{0.80} \\
&FDR~\cite{FDR}  &11.2 &5.19\sd{0.18}  &28.23\sd{0.46}  &11.2 &3.34\sd{0.16} &19.24\sd{0.07} \\
&DER++~\cite{DER}  &11.2 &9.56\sd{0.69}  &40.52\sd{0.47} &11.2  &10.30\sd{0.24} &29.20\sd{0.38} \\
&ERT~\cite{er_tricks}  &11.2 &9.95\sd{0.72}  &40.42\sd{1.57} &11.2 &10.28\sd{0.25} &28.53\sd{0.48} 
\\
& ASER$_\mu$~\cite{ASER} &11.2  &9.22\sd{0.25}  &41.09\sd{0.68} &11.2 &9.75\sd{0.57}  &31.71\sd{0.79}  
\\
&RM~\cite{rainbow}  &11.2 &8.39\sd{1.37}  &41.63\sd{0.74} &11.2 &8.53\sd{0.68} &28.30\sd{0.52} \\
&SCR~\cite{SCR} &11.2  &9.08\sd{0.74} &39.85\sd{0.93} &11.2 &8.81\sd{0.79} &29.62\sd{1.24}
\\
& \textbf{CVT (ours)} &\textbf{9.0}
&\textbf{14.71}\sd{1.04}
&\textbf{43.93}\sd{1.42}
&\textbf{9.4}  &\textbf{14.82}\sd{0.31} &\textbf{36.74}\sd{0.46}
 \\ \hline
 \multirow{8}{*}{1000} 
 &ER~\cite{ER}  &11.2 &12.46\sd{0.45}  &45.50\sd{0.61} &11.2 &10.42\sd{0.51} &34.26\sd{0.43} \\
 &AGEM~\cite{A-GEM}  &11.2 &4.92\sd{0.13}  &28.38\sd{0.15} &11.2 &3.66\sd{0.05} &23.56\sd{0.19}  \\
&iCaRL~\cite{iCaRL} &11.2 &6.91\sd{0.52}  &28.56\sd{0.37} &11.2 &8.93\sd{0.48} &22.37\sd{1.04} \\
&FDR~\cite{FDR}  &11.2 &5.27\sd{0.12}  &28.94\sd{0.36} &11.2 &3.58\sd{0.09} &21.28\sd{0.11} \\
&DER++~\cite{DER} &11.2 &12.97\sd{0.42}  &47.21\sd{0.33} &11.2 &13.94\sd{0.71} &40.02\sd{0.39} \\
&ERT~\cite{er_tricks} &11.2 &13.84\sd{0.77}  &44.65\sd{0.79} &11.2 &12.26\sd{0.23} &33.88\sd{0.60} \\
& ASER$_\mu$~\cite{ASER} &11.2  &12.26\sd{0.38}  &46.02\sd{0.82} &11.2 &11.38\sd{0.54}  &35.76\sd{1.28}  
\\
&RM~\cite{rainbow} &11.2 &11.73\sd{0.89}  &45.89\sd{0.64} &11.2 &11.85\sd{1.17} &32.72\sd{0.85} \\
&SCR~\cite{SCR} &11.2  &10.19\sd{1.14} &43.58\sd{1.20} &11.2 &10.74\sd{0.85} &31.84\sd{2.27} 
\\
& \textbf{CVT (ours)} &\textbf{9.0}
 &\textbf{16.54}\sd{1.22}
 &\textbf{48.50}\sd{0.88}
 &\textbf{9.4}  &\textbf{18.02}\sd{0.25} &\textbf{42.61}\sd{0.72}
\\ \hline
\end{tabular}  
}
\caption{
Results (overall accuracy~\%) on TinyImageNet and ImageNet100, which are averaged over three runs. \#Paras means the number of parameters in the model, which is counted by million.
}
\label{Otab:imagnet}
\end{table*}

\noindent\textbf{Metrics.}
We evaluate online CL methods in terms of accuracy and forgetting following~\cite{RW,A-GEM,DER}.
The accuracy is defined by $\mathbf{A}_T$=$\frac{1}{T} \sum_{t=1}^T a_{T,t}$, and the forgetting is defined by  $\mathbf{F}_T$=$\frac{1}{T-1} \sum_{t=1}^{T-1}{\text{max}_{i \in \{1,\dots, T-1\}}~(a_{i,t}-a_{T,t})}$, where $a_{T,t}$ is the inference accuracy on task $\T_t$ when the model finished learning task $\T_T$.

\noindent\textbf{Implementation Details.} 
In order to compare each method fairly, we train all networks using stochastic gradient descent (SGD) optimizer.
The images used for training are randomly cropped and flipped for each method following~\cite{DER,er_tricks,geodesic}.
We adopt 1 epoch with mini-batch size of 10 for all datasets, following~\cite{iCaRL,DER,er_tricks,SI}.
% For ImageNet100, we resize the images to 224$\times$224 and use batch-size of 128, step annealing learning-rate schedule ranged from 0.1 to 0.001, and the number of epochs of 100, which are used from~\cite{iCaRL,rainbow}.
Online continual learning baselines use ResNet18~\cite{resnet} as backbone and cross-entropy as the classification loss, following~\cite{A-GEM,geodesic,rainbow,DER,HAL,layerwise}.
The implementation of transformer block is based on LeViT~\cite{levit} and ViT~\cite{vit}. CVT framework employs GELU activation and dropout in transformer blocks and uses a global average pooling to the last activation map.

\subsection{Comparison to State-of-the-Art Methods}

\noindent\textbf{Evaluation on CIFAR100.}
Following the setting proposed in~\cite{iCaRL,DER_cvpr21}, we trains all 100 classes in several splits, including 10, 20 incremental tasks. Table~\ref{Otab:cifar100} summarizes the overall accuracy on CIFAR100 with 500 and 1000 memory sizes.
It is demonstrated that CVT outperforms other baselines by a considerable margin in different incremental splits, e.g., CVT can improve the accuracy of continual learning by more than 8\% in 10-split with 500 memory capacity. 
Especially in the case of small memory, the advantage of CVT is more obvious, which indicates CVT can effectively alleviate the imbalance issue in online CL. It is worth noting that although CVT uses fewer parameters (8.9M) than other methods (11.2M$\sim$33.8M), it can still achieve superior performance. One reason is that CVT inherits the merits of transformers for modeling the stream of tasks without stacking a lot of parameters, and besides, the number of parameters for the proposed learnable focuses is extremely small.

\begin{figure}[t]
    \centering
    \includegraphics[width=\textwidth]{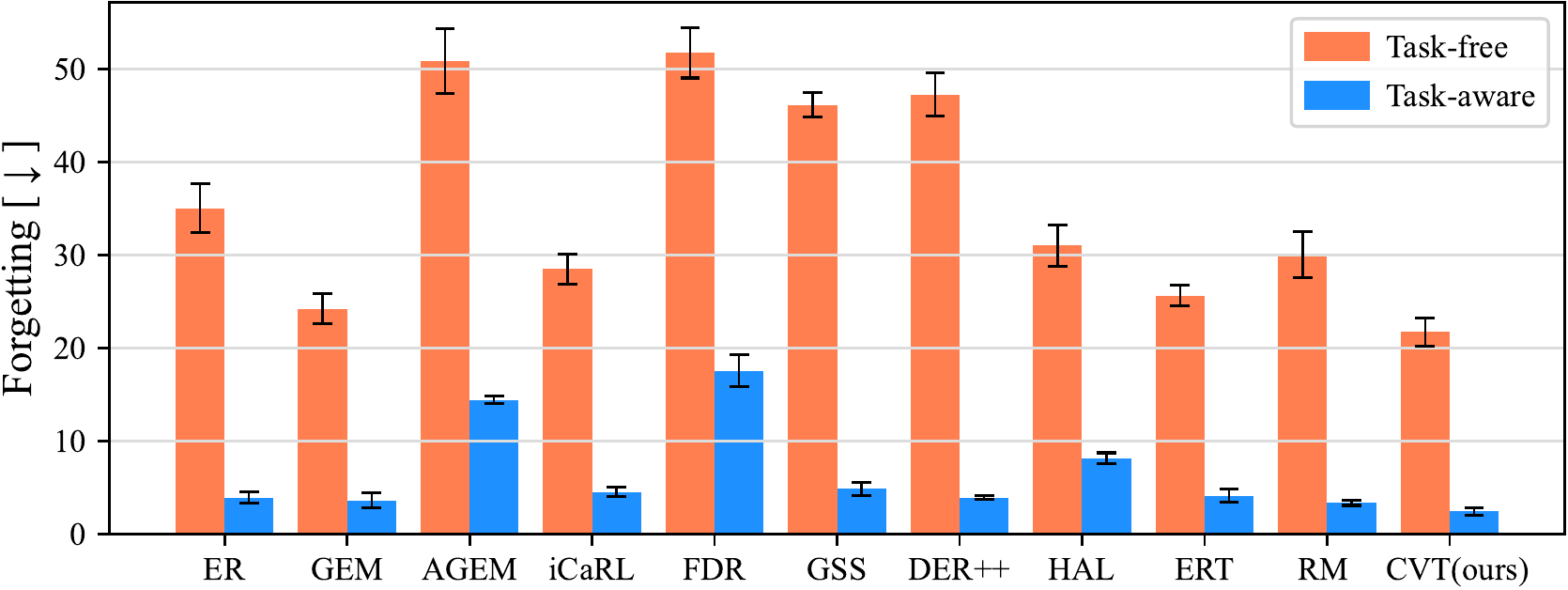}
    \caption{Forgetting results (\%) on CIFAR100 (lower is better).}
    \label{Ofig:forgettting}
\end{figure}

\noindent\textbf{Evaluation on ImageNet datasets.}
Table~\ref{Otab:imagnet} summarizes the evaluation results for the TinyImageNet and ImageNet100 datasets with 10 splits. It is demonstrated that CVT consistently surpasses other methods with a considerable margin for Task-free and Task-aware on TinyImageNet and ImageNet100 datasets.
Specifically, our method outperforms the state-of-the-art with about \textbf{4.3\%} for the Task-aware accuracy on the ImageNet100 benchmark.
For TinyImageNet benchmark, the Task-free accuracy is improved from 9.95\% to 14.71\%(\textbf{+4.76\%}). Moreover, CVT takes fewer parameters compared to other CNN-based methods.

\begin{figure}[t]
    \centering
    {
    \setlength\tabcolsep{15pt}
    \resizebox{\textwidth}{!}{
    \begin{tabular}{cc}
         \includegraphics[width=0.58\textwidth]{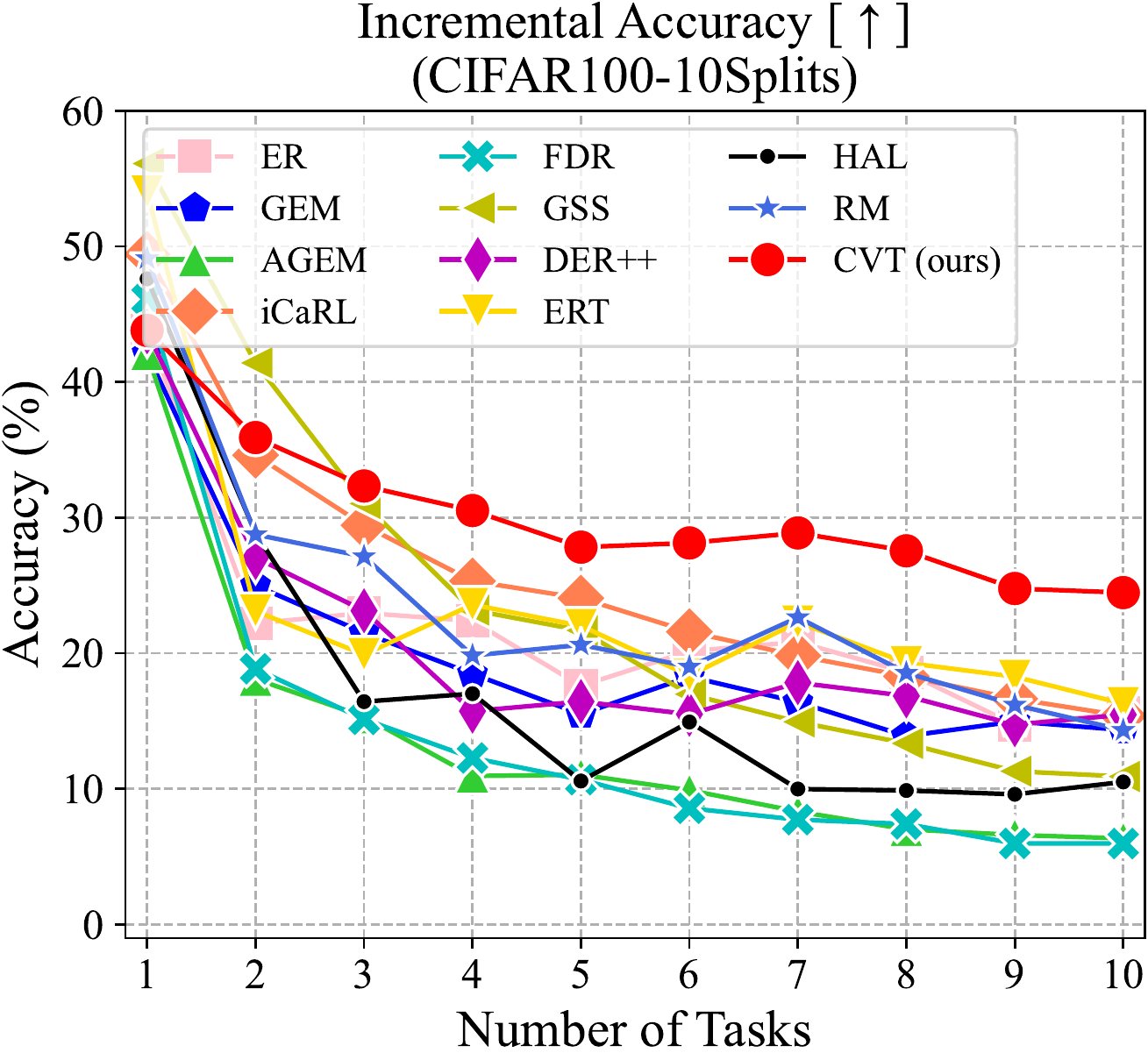} 
        &
        \includegraphics[width=0.58\textwidth]{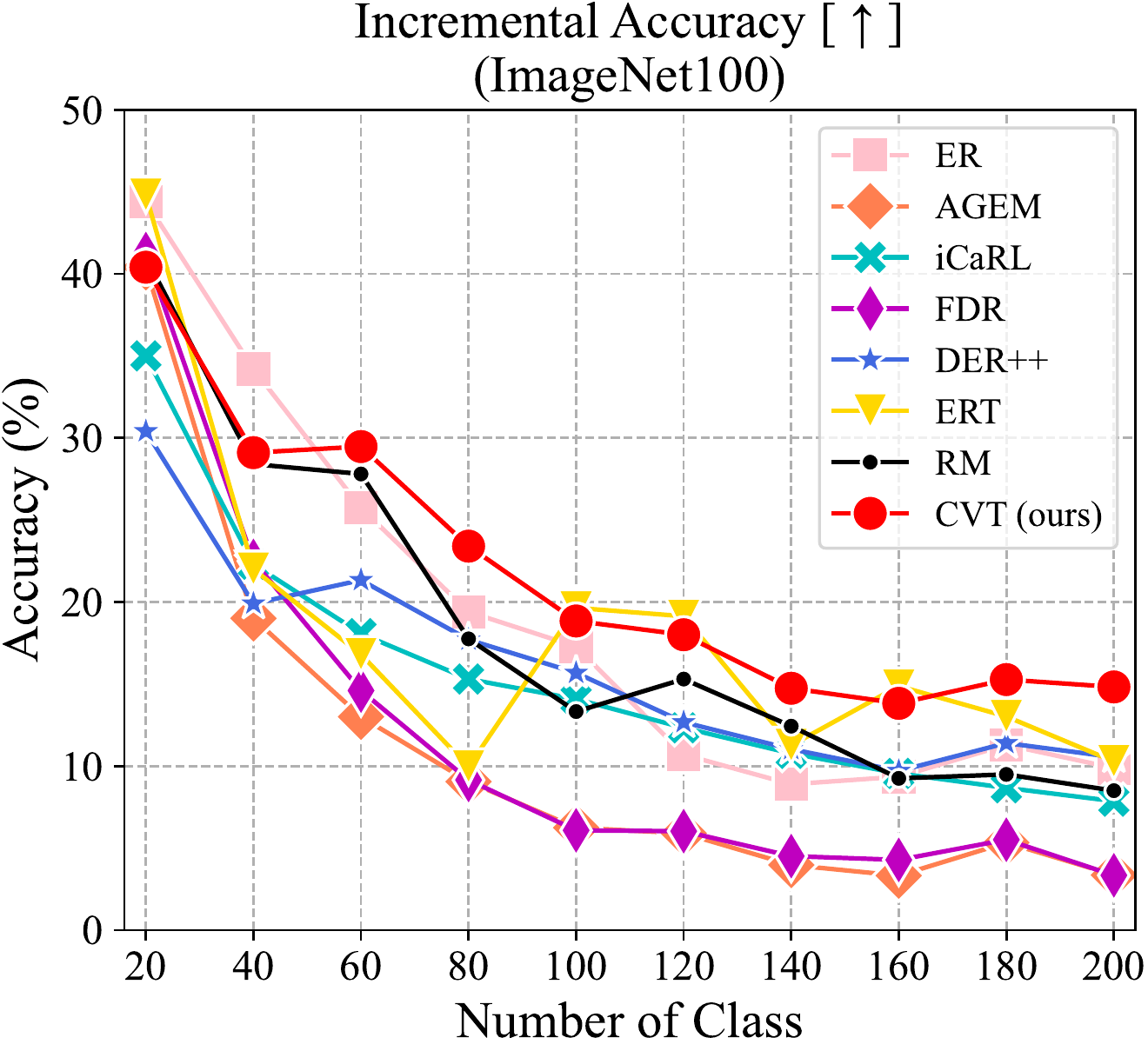} 
        \\
        \includegraphics[width=0.58\textwidth]{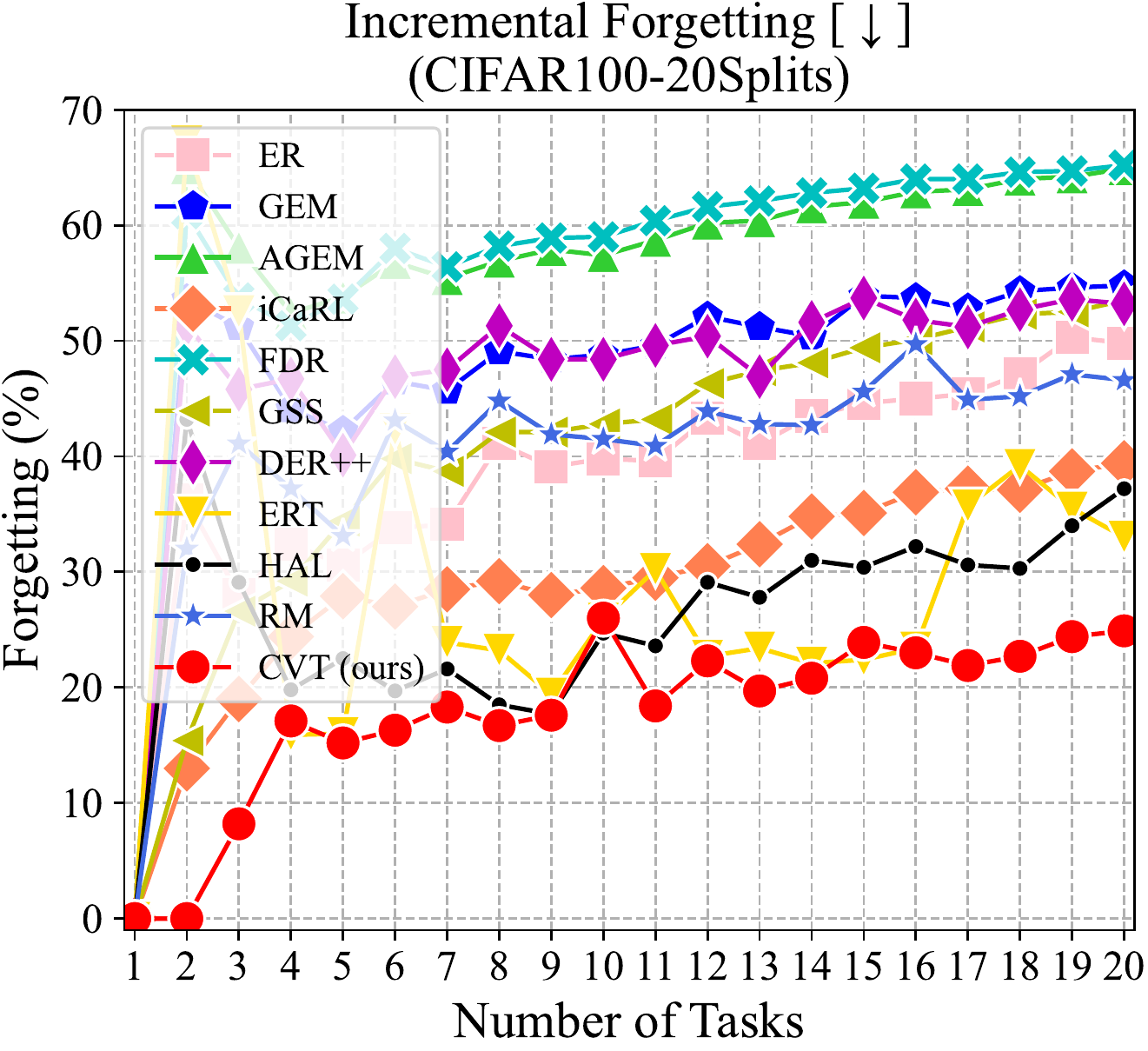} 
        &
        \includegraphics[width=0.58\textwidth]{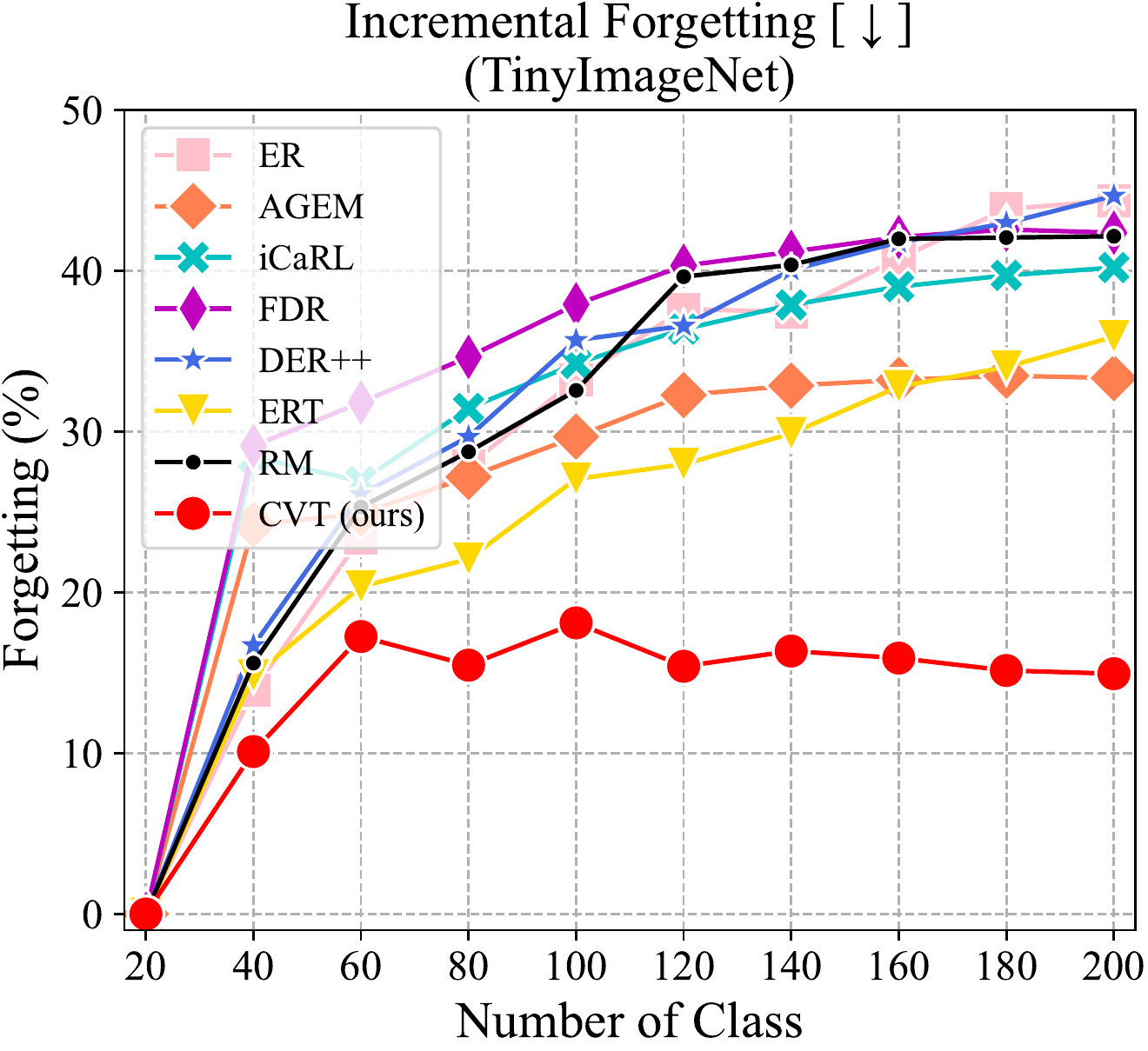} 
    \end{tabular}
    }
    }
    \caption{Incremental performance evaluated on tasks observed so far. [$\uparrow$] higher is better, [$\downarrow$] lower is better \textit{(best seen in color)}.  
    }
    \label{Ofig:plot_acc_forget}
\end{figure}

\noindent\textbf{Forgetting.}
To compare the alleviating forgetting capability, we assess the \textit{average forgetting}~\cite{DER,RW} that measures the performance degradation in subsequent tasks. As shown in \cref{Ofig:forgettting}, CVT suffers from less forgetting than all the other baselines in both of Task-free and Task-aware settings with memory buffer 1000 on CIFAR100. This is because CVT utilizes the focal contrastive loss and dual-classifier loss, which improve the stability of the vision transformer network.

\noindent\textbf{Incremental Performance.}
We also evaluate the \textit{average incremental performance}~\cite{iCaRL,DER} under the Task-free protocol with 500 memory buffer, which is the result of evaluating on all the tasks observed so far after completing each task. 
As illustrated in \cref{Ofig:plot_acc_forget}, the results are curves of accuracy and forgetting after each task. It is observed that during the learning process, most methods degrade rapidly as new tasks arrive, while our method consistently outperforms the state-of-the-art methods in both accuracy and forgetting.

\subsection{Ablation Study and Analysis}
\label{Osubsec:ablation}

\noindent\textbf{Comparison to Transformer and CNN Backbone.}
We compare CVT to Vision Transformer networks (ViT~\cite{vit}, LeViT~\cite{levit}, CoAT~\cite{CoAT}, and CCT~\cite{CCT}) and the CNN benchmark ResNet18~\cite{resnet} under the proposed rehearsal strategy in online continual learning.
Table~\ref{Otab:ablation} demonstrates the results of accuracy and forgetting on CIFAR100 and TinyImageNet with 500 memory.
It is observed that ViT is not up to the task of online continual learning, since it is “data-hungry” and only fits i.i.d.~large datasets. 
Besides, LeViT, CoAT, and CCT contain CNN structures to obtain inductive biases, which still suffer from catastrophic forgetting in online CL.
We can find that using Vision Transformer directly for online CL can not consistently outperform CNN-based networks.
Our proposed CVT architecture essentially inherits the merits of CNN and transformers and thus works well in online streaming data and modeling long-dependencies in the input data.
Moreover, CVT even takes fewer parameters to achieve better performance for online CL, which also benefits from the focal contrastive loss and the dual-classifier structure.

\noindent\textbf{Effect of Each Component.}
% We compare different proposed components for prevent forgetting previous knowledge.
To assess the effects of the components in CVT, we perform ablation study in terms of accuracy and forgetting.
From Table~\ref{Otab:ablation} we can observe that the proposed focal contrastive loss $\mL_\text{FC}$ plays an important role in alleviating catastrophic forgetting and accumulating knowledge.
However, if we simply use supervised contrastive learning loss $\mL_\text{SCL}$ to replace $\mL_\text{FC}$, we find that the forgetting problem is not mitigated compared to not using any contrastive loss. This is because using $\mL_\text{SCL}$ directly could cause a severe imbalance between new and past classes in online CL, which limits the learning of transferable representation.
While $\mL_\text{FC}$ can overcome the issue by utilizing the learnable focuses to boost the performance of online CL.
This supports that $\mL_\text{FC}$ can \textbf{rebalance contrastive learning between new and past classes and improve the inter-class distinction and intra-class aggregation}.
Besides, we can see the dual-classifier loss obtains 2.37\% and 7.97\% gain in terms of accuracy and forgetting on CIFAR100, respectively.
The results of Table~\ref{Otab:ablation} demonstrate the effectiveness of each component of CVT.

\begin{table}[t]
\centering
% \small
\setlength{\tabcolsep}{2pt}
\resizebox{\columnwidth}{!}{%{
\begin{tabular}{@{}lccc|ccc@{}}
\multirow{2}{*}{\textbf{Method}} & \multirow{2}{*}{$\#$\textbf{Paras}}  &\multicolumn{2}{c|}{\textbf{CIFAR100}}  &\multirow{2}{*}{$\#$\textbf{Paras}}  &\multicolumn{2}{c}{\textbf{TinyImageNet}}\\
& &Accuracy[$\uparrow$] &Forgetting[$\downarrow$] 
& &Accuracy[$\uparrow$] &Forgetting[$\downarrow$]  \\ 
\hline
ViT~\cite{vit}   &16.2 &8.48 &35.56 &16.3 &7.91  &42.26  \\
LeViT~\cite{levit} &10.9 &14.55 &44.53 &12.1 &9.02 &41.41
\\
% Twins~\cite{Twins} &24.1 & & & \\
CoAT~\cite{CoAT} &10.3 &13.17 &47.64 &11.3 &8.78  &39.80 \\
CCT~\cite{CCT} &\textbf{4.3} &13.86 &51.06 &\textbf{4.4} &9.97  & 40.21  
\\ \hline 
ResNet18~\cite{resnet} &11.2 &15.72 &43.82 &11.2 &9.56 &42.13
\\
ResNet18 + $\mL_\text{FC}$ &11.3 &17.49 &38.73 &11.3 &10.17  &36.49
\\ 
ResNet18 + dual-classifier &11.2 &18.84 &35.38 &11.2 &10.91  &33.55
\\ \hline
CVT - $\mL_\text{FC}$  &8.8 &19.92 &26.16 &8.9 &12.47 &19.43
\\
CVT - $\mL_\text{FC}$ + $\mL_\text{SCL}$  &8.9 &20.73 &25.52 &9.0 &12.59 &20.47
\\
CVT - dual-classifier   &8.9 &22.08 &29.81  &9.0 &13.63 &18.95
\\
CVT (ours) &8.9 &\textbf{24.45} &\textbf{21.86} &9.0 &\textbf{14.71} &\textbf{16.32}\\
\hline
\end{tabular}
}
\caption{Ablation study on backbone and each component of CVT.  ``-'' indicates the removal operation. ``+'' represents an add component operation. }
\label{Otab:ablation}
\end{table}

%% file: text/5conclu.tex
\section{Conclusion}
\label{Osec:conclu}
In this paper, we propose a novel attention-based framework, Contrastive Vision Transformer (CVT), to effectively mitigate the catastrophic forgetting for online CL. To the best of our knowledge, this paper is the first in the literature to design a Transformer for online CL. CVT contains external attention and learnable focuses to accumulate previous knowledge and maintain class-specific information. 
With a proposed focal contrastive loss in training, CVT rebalances contrastive continual learning between new and past classes and improves the inter-class distinction and intra-class aggregation.
Moreover, CVT designs a dual-classifier structure to decouple learning current classes and balancing all seen classes.
Extensive experimental results show that our approach significantly outperforms current state-of-the-art methods with fewer parameters. Ablation study validates the effectiveness of each proposed component.

\subsubsection{Acknowledgments}
This work is supported by ARC FL-170100117, DP-180103424, IC-190100031, and LE-200100049.